\PassOptionsToPackage{normalem}{ulem}
\ifdefined\XeTeXversion
  \def\pdfoutput=1{\let\pdfoutput\undefined}
  \def\pdfmapline#1{}
  \AddToHook{package/microtype/after}{%
    \renewcommand{\DisableLigatures}[2][]{}%
  }
\fi
\documentclass[]{bytedance_seed}

\ifdefined\XeTeXversion
  \usepackage[no-math]{fontspec}
  \newfontfamily\seedmedium{ByteSans-Medium.ttf}[
    Path=seed/,
    NFSSFamily=bytesansmedium,
    BoldFont=ByteSans-Bold.ttf]
\fi

\usepackage{amsmath,amsfonts,bm}

\def\eqref#1{equation~\ref{#1}}

\def\1{\bm{1}}

\DeclareMathAlphabet{\mathsfit}{\encodingdefault}{\sfdefault}{m}{sl}
\SetMathAlphabet{\mathsfit}{bold}{\encodingdefault}{\sfdefault}{bx}{n}

\usepackage{amsmath,amssymb,mathtools,amsthm}
\usepackage{booktabs}
\usepackage{multirow}
\usepackage{makecell}
\usepackage{array}
\usepackage{algorithm}
\usepackage{algpseudocode}
\usepackage{graphicx}
\usepackage{subcaption}
\usepackage{xcolor}
\usepackage{xspace}
\usepackage[normalem]{ulem}
\usepackage{microtype}
\usepackage{enumitem}
\usepackage{flafter}
\usepackage{placeins}
\usepackage{needspace}
\usepackage{hyperref}
\usepackage{url}

\titleformat*{\paragraph}{\normalfont\bfseries}

\newcommand{\ours}{\mbox{PISA}\xspace}
\newcommand{\meanrouter}{\mbox{PISA-1}\xspace}
\newcommand{\secondrouter}{\mbox{PISA-2}\xspace}
\newcommand{\flatbsa}{\mbox{BSA}\xspace}

\newcommand{\pisa}{{PISA}\xspace}

\title{Block Sparse Attention with Log-Linear Complexity}

\author[1,2,3]{Bohao Tang}
\author[3,\ddagger]{Zhen Qin}
\author[3]{Yuqi Pan}
\author[3]{Zheng Li}
\author[1,2,\dagger]{Pengfei Liu}

\affiliation[1]{Shanghai Jiao Tong University}
\affiliation[2]{Shanghai Innovation Institute}
\affiliation[3]{ByteDance Seed}

\contribution[\ddagger]{Project lead}
\contribution[\dagger]{Corresponding author}

\abstract{
Scaling language models to long contexts is limited by the quadratic cost of self-attention. Block sparse attention offers an efficient alternative, but selecting the retained blocks remains a bottleneck. Conventional block selection requires scoring all query–block pairs and therefore remains quadratic in sequence length.
To address this issue, we propose \pisa, a block-sparse attention mechanism that employs a pyramid Top-$K$ selection strategy. The main idea is to gradually narrow down the candidates across different levels, making it more efficient to find the most relevant keys. Specifically, we construct a coarse-to-fine hierarchy of keys and perform selection from the coarsest level. At each level, LogSumExp scoring is applied to a bounded candidate set to select candidates for the next finer level, continuing until the finest level is reached. Through pooling, we construct $O(\log N)$ levels of keys, yielding an overall complexity of $O(N\log N)$, where $N$ denotes the sequence length.
We develop hardware-aware Triton kernels for both training and inference, fusing hierarchical routing and LogSumExp scoring without materializing the query–key score matrix. We further evaluate our method on language modeling tasks. Compared with the baseline, our method achieves comparable performance on benchmarks such as commonsense reasoning while delivering better results on retrieval tasks.

}

\date{\today}
\correspondence{\email{tangbohao@sjtu.edu.cn}, \email{qinzhen.0109@bytedance.com}, \email{pengfei@sjtu.edu.cn}}

\begin{document}
\maketitle

\section{Introduction}
\label{sec:introduction}

Scaling language models to longer contexts has become increasingly important, but the computational complexity of full self-attention grows quadratically with sequence length, making it prohibitively expensive for long sequences~\citep{vaswani2017attention}. Sparse attention reduces this cost by restricting each query to a subset of keys~\citep{child2019sparse,beltagy2020longformer,zaheer2020bigbird}. Block-sparse attention groups keys and values into contiguous blocks of size $C$, allowing the sparse computation to use efficient tiled kernels~\citep{yuan2025nsa,jin2025moba,gao2024seerattention}. Block-sparse attention generally consists of two stages: Top-$K$ block selection and attention computation over the selected blocks. Once the Top-$K$ blocks are selected, attention is computed using the original keys and values within these blocks. Since each query attends to at most $KC$ keys, the second stage scales linearly with sequence length.

However, conventional Top-$K$ block selection computes a relevance score between each query and every key block, and then selects the $K$ highest-scoring blocks. For a sequence of length $N$ with a block size of $C$, there are approximately $N/C$ candidate key blocks. Since each query must scan all candidate key blocks, the selection stage still incurs a total complexity of $O(N^2/C)$ over the entire sequence. Therefore, although attention computation over the selected blocks scales linearly with sequence length, the Top-$K$ block-selection stage remains a major computational bottleneck in long-context modeling.

To address this bottleneck, we propose \textbf{\pisa} (\textbf{P}yram\textbf{i}d \textbf{S}parse \textbf{A}ttention), a block-sparse attention method based on hierarchical Top-$K$ block selection. \pisa first constructs a multilevel representation of key blocks through pooling. The finest level consists of fixed-size key blocks, while each coarser level is formed by aggregating adjacent representations from the level below. As the hierarchy becomes coarser, each block covers a larger portion of the sequence and the number of candidate blocks decreases, resulting in a coarse-to-fine pyramid of key representations.

Selection starts at the coarsest level of the hierarchy. At each level, \pisa{} computes LogSumExp (LSE) scores over a bounded set of candidate blocks and retains the Top-$K$ candidates. Each selected block is then expanded into its finer-grained sub-blocks at the next level, while the unselected blocks are discarded. This coarse-to-fine process continues until the finest level is reached. In this way, each query only evaluates a small number of promising blocks at each level, avoiding an exhaustive comparison with all fine-grained blocks.

The pooled key hierarchy contains $O(\log N)$ levels. With fixed Top-$K$ budget, branching factor, and leaf-block size, each query evaluates a bounded number of candidate blocks at every level, resulting in a per-query routing complexity of $O(\log N)$. Routing all queries in a sequence of length $N$ therefore yields an overall block-selection complexity of $O(N\log N)$.

To execute this process efficiently on modern hardware, we develop hardware-aware Triton kernels for both training and inference~\citep{tillet2019triton}. Our implementation fuses multiple levels of hierarchical routing and LSE scoring into a single kernel, allowing intermediate candidates to be directly expanded and filtered at the next level without repeatedly reading and writing intermediate results across routing levels. The kernel maintains only a bounded candidate set at the current level and does not materialize or store a dense query--key score matrix, thereby reducing memory traffic and intermediate-tensor overhead for long sequences.

We evaluate our method on language modeling tasks. Compared with the conventional block-sparse attention baseline \flatbsa, our method achieves comparable performance on commonsense reasoning benchmarks while delivering better results on retrieval tasks. These results indicate that coarse-to-fine candidate search reduces selection complexity while preserving blocks relevant to downstream modeling. Top-$K$ selection benchmarks with sequence lengths of up to 256K tokens further show that hierarchical routing is more efficient than conventional single-level selection for long sequences.

\section{Related Work}
\label{sec:related_work}

\begin{table}[htbp]
\caption{Training settings and computational complexity of selected
sparse attention methods. $N$ denotes the sequence length.
Complexities include selection and attention computation.
$\checkmark$ denotes trainable, and $\times$ denotes a
training-free method.
For LLSA~\citep{zhou2025llsa}, which is designed for diffusion models, the listed cost is per
attention pass rather than autoregressive prefill.}
\label{tab:sparse_attention_comparison}
\centering
\footnotesize
\setlength{\tabcolsep}{4pt}
\renewcommand{\arraystretch}{1.0}
\begin{tabular*}{\textwidth}{@{\extracolsep{\fill}}lccc@{}}
\toprule
\makecell[c]{Method} & \makecell[c]{Trainable}
& \makecell[c]{Prefill complexity} & \makecell[c]{Decode complexity} \\
\midrule
HiP~\citep{zhang2024hip} & $\times$ & $O(N\log N)$ & $O(\log N)$ \\
HISA~\citep{xu2026hisa} & $\times$ & $O(N^2)$ & $O(N)$ \\
MoBA~\citep{jin2025moba} & $\checkmark$ & $O(N^2)$ & $O(N)$ \\
NSA~\citep{yuan2025nsa} & $\checkmark$ & $O(N^2)$ & $O(N)$ \\
HiLS~\citep{hu2026hils} & $\checkmark$ & $O(N^2)$ & $O(N)$ \\
LLSA~\citep{zhou2025llsa} & $\checkmark$ & $O(N\log N)$ & --- \\
\midrule
\flatbsa~\citep{yuan2025nsa} & $\checkmark$ & $O(N^2)$ & $O(N)$ \\
\textbf{\pisa} (ours) & $\checkmark$ & $O(N\log N)$ & $O(\log N)$ \\
\bottomrule
\end{tabular*}
\end{table}

Sparse attention reduces attention computation by limiting the keys
and values that each query attends to.
Existing methods either apply sparsity to pretrained models without
updating their weights or incorporate sparse attention during model
training. We review these training-free and trainable methods below.
Table~\ref{tab:sparse_attention_comparison} summarizes the training settings
and computational complexity of selected methods closely related to ours.

\paragraph{\upshape\bfseries Training-Free Sparse Attention.}
Training-free sparse attention reduces inference cost without retraining
the model. During prefill, methods use head-specific sparse patterns
~\citep{jiang2024minference} or estimate block importance
through sampling and online filtering
~\citep{xu2025xattention,zhang2025spargeattention}.
During decoding, methods reduce KV-cache accesses by selecting relevant
tokens or pages before loading their full keys and values.
Selection scores are computed from a subset of key dimensions
~\citep{ribar2024sparq} or page-level minimum and maximum key values
~\citep{tang2024quest}.
To reduce the number of token scores, two-stage methods first select
blocks or clusters using their summaries, then score tokens only within
the selected groups~\citep{xu2026hisa,ni2026doublep}.
HiP~\citep{zhang2024hip} repeats selection across multiple levels:
it divides retained regions into smaller ones and keeps a fixed number
at each step, giving log-linear complexity for a fixed selection budget.

\paragraph{\upshape\bfseries Trainable Sparse Attention.}
Trainable methods train or adapt models with sparse attention.
Early methods combine local attention with selected long-range connections
~\citep{child2019sparse,beltagy2020longformer,zaheer2020bigbird}.
Query-dependent methods such as MoBA~\citep{jin2025moba} and
InfLLM-V2~\citep{zhao2025infllmv2} use pooled block summaries, while
NSA~\citep{yuan2025nsa} and HiLS~\citep{hu2026hils} use learned block
summaries to select relevant blocks.
Other methods learn gates or use differentiable rules for block selection
~\citep{gao2024seerattention,huang2026dashattention}.
IndexCache~\citep{bai2026indexcache} and LongCat~\citep{zan2026longcat}
use cross-layer distillation to enable index reuse across layers.
SSA and H-SSA~\citep{mao2026ssa} use continued pretraining (CPT)
to learn single-level and hierarchical gist summaries, respectively,
for chunk selection.
LLSA~\citep{zhou2025llsa} performs hierarchical Top-$K$ selection for
diffusion transformers and includes both original and compressed keys
and values in the final attention computation.

\section{Method}
\label{sec:method}

\begin{figure}[t]
    \centering
    \includegraphics[width=0.9\linewidth,pagebox=cropbox]{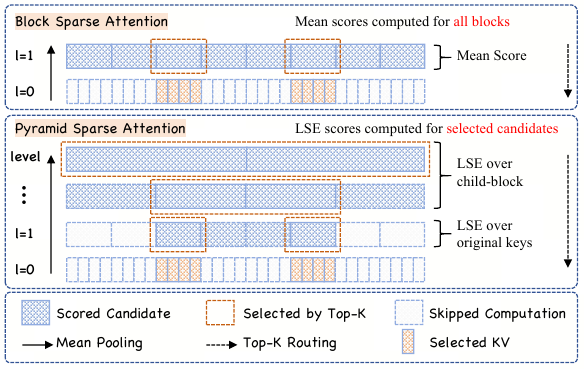}
    \caption{Comparison of two different Top-$K$ selection methods for sparse attention. Top: BSA computes a score for each key block before performing Top-$K$ selection. Bottom: PISA constructs a fine-to-coarse hierarchy of key blocks through mean pooling and expands only the selected candidates from coarse to fine during selection. Intermediate blocks are scored using LSE over their child summaries, while leaf blocks are scored using exact LSE over their original keys.}
    \label{fig:method_overview}
\end{figure}

\subsection{Block Sparse Attention}
\label{sec:block_sparse_attention}

Block sparse attention (BSA)~\citep{yuan2025nsa,jin2025moba} partitions the key sequence into contiguous
blocks and selects a subset of these blocks for each query. The values at the
same positions are selected together with each key block. 

Let
$q_t,k_j,v_j\in\mathbb R^d$ denote the query at position $t$ and the key and
value at position $j$. For block size $C$, the $i$-th key block is
\begin{equation}
K_i=\left[k_{(i-1)C+1},\ldots,k_{\min(iC,N)}\right],
\label{eq:bsa_key_block}
\end{equation}
where
$i\in\{1,\ldots,M\}$ and $M=\lceil N/C\rceil$. BSA first applies a block
mapping $f$, such as mean pooling~\citep{jin2025moba} or a learned linear projection~\citep{yuan2025nsa}, to obtain a
summary vector
\begin{equation}
\bar{k}_i=f(K_i)\in\mathbb R^d.
\label{eq:bsa_summary}
\end{equation}

The summary vector is used for selection, for query $q_t$, BSA computes one score for every block:
\begingroup
\begin{align}
 \mathbf{s}_t
 &=\left[s_{t,1},\ldots,s_{t,M}\right],\\
 s_{t,i}&=\frac{q_t^\top\bar{k}_i}{\sqrt d}.
 \label{eq:bsa_block_scores}
\end{align}
\endgroup
BSA retains the indices of the $K$ largest scores,
\begin{equation}
\mathcal I_t=\operatorname{TopK}_K(\mathbf{s}_t).
\label{eq:bsa_topk}
\end{equation}

Let $\mathcal T_t=\bigcup_{i\in\mathcal I_t}
\{(i-1)C+1,\ldots,\min(iC,N)\}$ contain the token positions covered by the
selected blocks. The attention output is then computed from the original keys
and values at these positions:
\begingroup
\begin{equation}
 o_t
 =
\frac{
   \displaystyle\sum_{j\in\mathcal T_t}
   \exp\!\left(q_t^\top k_j/\sqrt d\right)v_j
 }{
   \displaystyle\sum_{j\in\mathcal T_t}
   \exp\!\left(q_t^\top k_j/\sqrt d\right)
 }.
 \label{eq:selected_sparse_attention}
\end{equation}
\endgroup
The final attention reads the
original keys and values from the selected blocks. It therefore accesses at
most $KC$ keys per query, but scoring all $M=\lceil N/C\rceil$ block summaries
for every query still costs $O(N^2d/C)$.

\subsection{Pyramid Top-$K$ Block Selection}
\label{sec:aggregation}
\label{sec:routing_algorithm}

\pisa performs hierarchical Top-$K$ block selection on a key-block hierarchy
constructed from fine to coarse. Selection proceeds in the opposite direction and expands
only the blocks retained at the previous level. We first define the hierarchy
and candidate sets, and then introduce the LSE score used for selection.

\paragraph{Fine-to-coarse block aggregation.}
We first construct the key-block hierarchy from fine to coarse. At level
$\ell\geq0$, let $M_\ell$ denote the number of blocks and $K_i^{(\ell)}$
the $i$-th block, with $i\in\{1,\ldots,M_\ell\}$. Level $0$ contains the
original keys, each represented as a singleton block:
\begin{align}
 M_0&=N,\\
 K_i^{(0)}&=[k_i].
\end{align}

Let $g_\ell$ be the maximum number of adjacent level-$\ell$ blocks combined
into one level-$(\ell+1)$ block. We use $g_0=C$ and, for $\ell\geq1$,
\begin{equation}
 g_\ell=g=2.
\end{equation}
Let $\operatorname{Ch}_\ell(i)$ denote the indices of the
level-$\ell$ children of $K_i^{(\ell+1)}$. For $\ell\geq0$, the hierarchy
is constructed as
\begingroup
\begin{align}
 M_{\ell+1}&=\left\lceil\frac{M_\ell}{g_\ell}\right\rceil,\\
 \operatorname{Ch}_\ell(i)
 &=\left\{(i-1)g_\ell+1,\ldots,\min(ig_\ell,M_\ell)\right\},\\
 K_i^{(\ell+1)}
 &=\operatorname{Concat}\limits_{r\in\operatorname{Ch}_\ell(i)}
 K_r^{(\ell)}.
 \label{eq:key_hierarchy}
\end{align}
\endgroup
Here, $r$ indexes the child blocks, and $\operatorname{Concat}$ preserves
their sequence order. The first aggregation step forms the leaf blocks,
with
\begin{align}
 M_1&=M=\lceil N/C\rceil,\\
 K_i^{(1)}&=K_i.
\end{align}

Each block $K_i^{(\ell)}$ has a summary vector
$\bar{k}_i^{(\ell)}\in\mathbb R^d$. Starting from
$\bar{k}_i^{(0)}=k_i$, we compute the summaries recursively by mean pooling:
\begingroup
\begin{equation}
 \bar{k}_i^{(\ell+1)}
 =
 \frac{1}{|\operatorname{Ch}_\ell(i)|}
 \sum_{r\in\operatorname{Ch}_\ell(i)}
 \bar{k}_r^{(\ell)},
 \qquad \ell\geq0.
 \label{eq:mean_pyramid}
\end{equation}
\endgroup
The denominator counts the valid child blocks. When child blocks are equally
sized at every aggregation step, the summary equals the mean of all original
keys in the block. Queries remain token-wise throughout selection.

\paragraph{Coarse-to-fine block selection.}
Selection proceeds from the coarsest level $L$ to the leaf-block level $1$.
For each query $q_t$, we initialize the candidate index set at the coarsest
level as $\mathcal{A}_t^{(L)}=\{1\}$.

At level $\ell$, $\mathcal{A}_t^{(\ell)}$ denotes the candidate index set,
$s_{t,i}^{(\ell)}$ denotes the score of candidate block $i$, and
$\mathbf{s}_t^{(\ell)}$ collects the scores of all candidate blocks.
The operator $\operatorname{Select}_K$ returns the indices of the $K$
highest-scoring blocks, denoted by $\mathcal{I}_t^{(\ell)}$. The candidate
set at level $\ell-1$ is then constructed from the children of the retained
blocks in $\mathcal{I}_t^{(\ell)}$. The complete selection and expansion
process is defined as
\begingroup
\begin{align}
    \mathbf{s}_t^{(\ell)}
    &=
    \left[s_{t,i}^{(\ell)}\right]_{i\in\mathcal{A}_t^{(\ell)}},\\
    \mathcal{I}_t^{(\ell)}
    &=
    \operatorname{Select}_K
    \left(
        \mathcal{A}_t^{(\ell)},
        \mathbf{s}_t^{(\ell)}
    \right),\\
    \mathcal{A}_t^{(\ell-1)}
    &=
    \bigcup_{i\in\mathcal{I}_t^{(\ell)}}
    \operatorname{Ch}_{\ell-1}(i),
    \quad (\ell>1).
    \label{eq:hierarchical_topk}
\end{align}
\endgroup
If a level contains at most $K$ candidate blocks, all candidates are retained
without scoring. Otherwise, $\operatorname{Select}_K$ retains at most $K$
blocks. For $\ell>1$, each retained block has at most $g_{\ell-1}=g$
children, where we set $g=2$ in practice. Therefore,
\begingroup
\begin{align}
    \left|\mathcal{I}_t^{(\ell)}\right|
    &\leq K,\\
    \left|\mathcal{A}_t^{(\ell-1)}\right|
    &\leq
    g\left|\mathcal{I}_t^{(\ell)}\right|
    \leq gK.
    \label{eq:candidate_bound}
\end{align}
\endgroup
The final set $\mathcal{I}_t=\mathcal{I}_t^{(1)}$ identifies the original
key and value blocks used in the sparse attention computation. Appendix~\ref{app:causal_ablation} gives the complete
forced-block policy.

\paragraph{LogSumExp block scoring.}

For query $q_t$, \pisa assigns candidate $K_i^{(\ell)}$ the LSE score
\begingroup
\begin{equation}
 s_{t,i}^{(\ell)}
 =
 \log\sum_{r\in\operatorname{Ch}_{\ell-1}(i)}
 \exp\!\left(\frac{q_t^\top\bar{k}_r^{(\ell-1)}}{\sqrt d}\right),
 \qquad \ell\geq1.
 \label{eq:direct_child_lse}
\end{equation}
\endgroup
Each score reduces at most $g_{\ell-1}$ logits: $C$ at the leaf level and
$g$ at intermediate levels. Since $\bar{k}_r^{(0)}=k_r$, a completed
historical leaf receives its exact raw-key LSE score, whereas an intermediate
block is scored from its child summaries. Appendix~\ref{app:scoring_details}
gives further scoring details.

\subsection{Kernel Implementation}
\label{sec:kernel}
\label{sec:fused_kernel}

Following NSA~\citep{yuan2025nsa}, query heads within the same GQA group
share the same selected key blocks. We design different kernels for
training/prefill and decoding. For training and prefill, we adopt a
two-stage kernel design to reduce IO overhead. The first stage performs
hierarchical selection over the intermediate levels, while the second
stage scores the original key blocks. For decoding, we use a
single-stage kernel to avoid the overhead of multiple kernel launches.

\begin{algorithm}[h]
\caption{Two-stage block selection for PISA}
\label{alg:kernel_schedules}
\begin{algorithmic}[1]
\Statex \textbf{Input:}
query vectors $\{q_{t,h'}\}$,
key summaries
$\{\bar{k}_{i,h}^{(\ell)}\}_{\ell=1}^{L}$,
original key blocks $\{K_{i,h}^{(1)}\}$,
initial candidate sets $\{\mathcal{A}_{t,h}^{(L)}\}$,
block budget $K$, and maximum query positions per program $Q_{\mathrm{tile}}$

\Statex \textbf{Output:}
selected leaf-block indices $\{\mathcal{I}_{t,h}^{(1)}\}$

\Statex \textit{Stage 1: Intermediate-level selection}

\ForAll{query positions $t$ and KV heads $h$ in parallel}
    \State load $\{q_{t,h'}:h'\in\mathcal{H}(h)\}$ once;
    initialize
    $\mathcal{A}\gets\mathcal{A}_{t,h}^{(L)}$

    \For{$\ell=L,L-1,\ldots,2$}
        \ForAll{$i\in\mathcal{A}$ requiring scores}
            \State
            $s_{h',t,i}^{(\ell)}\gets
            \log\sum_{r\in\operatorname{Ch}_{\ell-1}(i)}
            \exp\!\left(q_{t,h'}^\top\bar k_{r,h}^{(\ell-1)}/\sqrt d\right)$
            \State
            $u_{h,t,i}^{(\ell)}
            \gets
            \displaystyle\sum_{h'\in\mathcal{H}(h)}
            s_{h',t,i}^{(\ell)}$
        \EndFor

        \State select
        $\mathcal{I}\gets
        \operatorname{Select}_K
        \left(
            \mathcal{A},
            \{u_{h,t,i}^{(\ell)}\}_{i\in\mathcal{A}}
        \right)$

        \State expand
        $\mathcal{A}\gets
        \displaystyle\bigcup_{i\in\mathcal{I}}
        \operatorname{Ch}_{\ell-1}(i)$,
        where $\operatorname{Ch}_{\ell-1}(i)$ indexes block $i$'s children
    \EndFor

    \State store
    $\mathcal{A}_{t,h}^{(1)}\gets\mathcal{A}$
\EndFor

\Statex \textit{Stage 2: Original key-block scoring}

\State For each KV head $h$ and leaf block $i$, collect the query positions $t$
that require a score for $i\in\mathcal A_{t,h}^{(1)}$,
divide each query cluster into disjoint groups $I_{h,i,j}$ with
$|I_{h,i,j}|\le Q_{\mathrm{tile}}$. 
\ForAll{groups $I_{h,i,j}$ in parallel}
    \State load $K_{i,h}^{(1)}$ once and
    $\{q_{t,h'}:t\in I_{h,i,j},\ h'\in\mathcal H(h)\}$
    \State $s_{h',t,i}^{(1)}\gets
    \log\sum_{r\in\operatorname{Ch}_0(i)}
    \exp\!\left(q_{t,h'}^\top k_{r,h}/\sqrt d\right)$
    \State store $u_{h,t,i}^{(1)}\gets\sum_{h'\in\mathcal H(h)}s_{h',t,i}^{(1)}$
    for each query position $t\in I_{h,i,j}$
\EndFor

\Statex \textit{Final Top-$K$ selection}

\ForAll{query positions $t$ and KV heads $h$ in parallel}
    \State
    $\mathcal{I}_{t,h}^{(1)}
    \gets
    \operatorname{Select}_K
    \left(
        \mathcal{A}_{t,h}^{(1)},
        \{u_{h,t,i}^{(1)}\}_{i\in\mathcal{A}_{t,h}^{(1)}}
    \right)$
\EndFor

\State \Return $\{\mathcal{I}_{t,h}^{(1)}\}$
\end{algorithmic}
\end{algorithm}

\paragraph{Training/Prefill Stage 1: Intermediate-level selection.}

In the first stage, we parallelize over the query position $t$ and KV
head $h$. Let $\mathcal{H}(h)$ denote the set of query heads that share
KV head $h$. Each kernel program loads the corresponding query vectors
once and sequentially processes the intermediate levels from $\ell=L$
to $\ell=2$.

At level $\ell$, $\mathcal{A}_{t,h}^{(\ell)}$ denotes the candidate block
indices. Using the child summaries of each candidate block, the program
computes an LSE score for every candidate and query head.
It then sums
the scores over all query heads sharing KV head $h$:
\begingroup
\begin{equation}
    u_{h,t,i}^{(\ell)}
    =
    \sum_{h'\in\mathcal{H}(h)}
    s_{h',t,i}^{(\ell)}.
    \label{eq:kernel_grouped_score}
\end{equation}
\endgroup
Here, $s_{h',t,i}^{(\ell)}$ is the LSE score of candidate block
$i\in\mathcal{A}_{t,h}^{(\ell)}$ for query head $h'$. The program applies
$\operatorname{Select}_K$ to obtain the retained indices
$\mathcal{I}_{t,h}^{(\ell)}$ and expands the selected blocks into their
children to form $\mathcal{A}_{t,h}^{(\ell-1)}$. The candidate indices
remain local to the program throughout this loop. If at most $K$ candidates are valid, all are retained without scoring.

\paragraph{Training/Prefill Stage 2: Original key-block selection.}

After the first stage, each query has at most $gK$ candidate key blocks.
For each KV head, we group queries that need to score the same
candidate key block into a query cluster. Each cluster is divided into
tiles of at most $Q_{\mathrm{tile}}$ queries, and tiles across all clusters
are processed in parallel. Each kernel program loads the corresponding
key block once and reuses it to jointly score the queries in its tile.
In our implementation, we set $Q_{\mathrm{tile}}=4$.

Let $G_Q = |\mathcal{H}(h)|$ denote the number of query heads sharing the same KV head.
Each program loads a key tile of shape $d\times C$ covering one
leaf block and a query tile of shape $(Q_{\mathrm{tile}}G_Q)\times d$.
The program computes
\begingroup
\begin{equation}
    \underbrace{(Q_{\mathrm{tile}}G_Q)\times d}_{\text{query tile}}
    \;\times\;
    \underbrace{d\times C}_{\text{key tile}}
    \;\longrightarrow\;
    (Q_{\mathrm{tile}}G_Q)\times C.
\end{equation}
\endgroup
The program applies LSE over keys for each query head, then sums across the $G_Q$ query heads:
\begin{equation}
    (Q_{\mathrm{tile}}G_Q)\times C
    \;\xrightarrow{\operatorname{LSE}}\;
    Q_{\mathrm{tile}}G_Q
    \;\xrightarrow{\operatorname{sum\ over\ heads}}\;
    Q_{\mathrm{tile}}.
\end{equation}
After the scores are stored, a final kernel independently applies
$\operatorname{Select}_K$ for each query and KV head.

\paragraph{Decoding.}

During decoding, we cache the mean pyramid and update only the
current leaf mean and its ancestor path for each new token.
Block selection uses a single kernel that follows Stage~1 but continues
to $\ell=1$, directly scoring the original key blocks.
Combining all selection levels into a single
kernel avoids the overhead of launching multiple kernels for each decoding step.

\subsection{Complexity}
\label{sec:complexity}

\paragraph{Computational complexity.}

Because the pyramid key hierarchy is constructed through downsampling,
it contains $L=O(\log N)$ levels. At each level, \pisa scores at most $gK$
candidate blocks for each
query. Each intermediate candidate uses at most $g$ child
summaries, and each leaf candidate uses at most $C$ original keys.
With fixed $g$, $K$, block size, head counts, and head dimension, the selection
complexity for a query is $O(\log N)$. Processing all $N$ queries during training or prefill costs $O(N\log N)$.
Decoding costs $O(\log N)$ per step on average, including the
occasional cost of expanding the cache.

\paragraph{Why do training/prefill and decoding use different kernel designs?}

A natural question is why Stage~1 does not continue directly to
$\ell=1$ during training and prefill. We compare the two designs' leaf-level Q/K IO costs for one KV head,
excluding the initial query load shared by both.

In the two-stage design, each query is associated with at most $gK$
candidate key blocks. Loading the query vectors for all query--block
pairs therefore incurs an IO cost of $O(NgKG_Qd)$.
Since query tiles are processed in parallel, each key block is loaded
\begingroup
\begin{equation}
O\!\left(\frac{NgK}{(N/C)Q_{\mathrm{tile}}}\right)
=O\!\left(\frac{CgK}{Q_{\mathrm{tile}}}\right)
\end{equation}
\endgroup
times on average.
Since there are $N/C$ key blocks and loading one key block costs
$O(Cd)$, the total key IO cost is
\begingroup
\begin{equation}
O\!\left(\frac{CgK}{Q_{\mathrm{tile}}}\cdot\frac{N}{C}\cdot Cd\right)
=O\!\left(\frac{NCgKd}{Q_{\mathrm{tile}}}\right).
\end{equation}
\endgroup
Under this approximation, the two-stage leaf-level Q/K IO cost is
\begingroup
\begin{equation}
    O\!\left(
        NgKd
        \left(
            G_Q+\frac{C}{Q_{\mathrm{tile}}}
        \right)
    \right).
    \label{eq:two_stage_prefill_io}
\end{equation}
\endgroup

In comparison, a single-stage kernel would load at most $gK$ original
key blocks per query at $\ell=1$. With $C$ keys per block, its additional
leaf-level IO cost is 
\begingroup
\begin{equation}
    O(NgKCd).
    \label{eq:stage2_prefill}
\end{equation}
\endgroup
When $G_Q+\frac{C}{Q_{\mathrm{tile}}}<C$,
the two-stage design reduces the leaf-level Q/K IO cost by
reusing each loaded key block across multiple queries.
In our implementation, we set $G_Q=16$, $Q_{\mathrm{tile}}=4$, and $C=64$,
which gives
\begingroup
\begin{equation}
    G_Q+\frac{C}{Q_{\mathrm{tile}}}
    =16+\frac{64}{4}=32<64=C.
\end{equation}
\endgroup
Thus, under these assumptions, the two-stage design has
a lower leaf-level Q/K IO cost for our configuration.

During decoding, each cluster contains only one query, so there is no
cross-query key reuse. A two-stage kernel would load $gKC$ keys and
load the query vectors $gK$ times at $\ell=1$, giving an IO cost of 
\begingroup
\begin{equation}
O\!\left(gK(C+G_Q)d\right).
\end{equation}
\endgroup
In contrast, the single-stage kernel reuses the query
vectors already loaded during intermediate-level selection.
Its additional leaf-level IO cost is 
\begingroup
\begin{equation}
O(gKCd).
\end{equation}
\endgroup

The fused selector avoids reloading queries at the leaf level
and requires no separate grouping or final-selection launches.
Accordingly, we use the two-stage kernel for training and prefill and
the single-stage kernel for decoding.

\section{Experiments}
\label{sec:experiments}

\subsection{Experimental setup}
\label{sec:experimental_setup}

We compare \ours with Full Attention, BSA, NSA~\citep{yuan2025nsa}, and
HiLS~\citep{hu2026hils} at the 418M, 1.47B, and 2.67B backbone scales,
using the same decoder-only backbone configuration at each scale and matched
training settings. BSA uses the selected-attention branch of NSA with
mean-pooled key summaries. NSA
retains its learned compression and selected-attention branches, while
HiLS retains its learnable landmark-based routing.
We pretrain the models for 100B tokens at a
sequence length of 4K, followed by 10B additional tokens of continued
pretraining (CPT) at 16K. Sparse methods use block size $C=64$ throughout,
with block budgets $K=8$ and $K=32$ for the two stages, respectively.
We report training loss, language-model perplexity, multiple-choice accuracy,
and containment accuracy, and use RULER~\citep{hsieh2024ruler} to measure
long-context retrieval. Complete training and evaluation protocols are
given in Appendix~\ref{app:setup}.

\subsection{Language Modeling and Downstream Evaluation}
\label{sec:main_results}
\label{sec:native_results}

Table~\ref{tab:native_main} compares language-modeling and downstream
performance at three model scales. \ours performs comparably to BSA, NSA,
and HiLS on language modeling and commonsense reasoning. Across the six
containment tasks, \ours achieves the highest average accuracy among sparse
methods at all three scales, although Full Attention retains a higher average.

\definecolor{avgshade}{RGB}{241,244,248}
\newcommand{\avgcell}[1]{\begingroup\setlength{\fboxsep}{1pt}\colorbox{avgshade}{\strut #1}\endgroup}
\newcommand{\metrichead}[2]{\shortstack{#1\\#2}}

\begin{table*}[htb]
\caption{Language modeling and downstream evaluation after 100B tokens of
pretraining at 4K. Baselines include NSA~\citep{yuan2025nsa} and
HiLS~\citep{hu2026hils}. Loss is the final training loss, and accuracies are
reported as percentages. Gray columns show group averages; boldface marks
the best sparse loss and group averages at each scale.}
\label{tab:native_main}
\centering
\scriptsize
\setlength{\tabcolsep}{1.5pt}
\resizebox{\textwidth}{!}{%
\begin{tabular}{@{}lc|c|ccc|ccccccccc|ccccccc@{}}
\toprule
\multirow{2}{*}[-3ex]{Method}
& \multirow{2}{*}[-3ex]{Params}
& \multirow{2}{*}[-3ex]{\metrichead{Loss}{$\downarrow$}}
& \multicolumn{3}{c|}{Perplexity}
& \multicolumn{9}{c|}{Multiple-choice}
& \multicolumn{7}{c}{Containment} \\
\cmidrule(lr){4-6}\cmidrule(lr){7-15}\cmidrule(lr){16-22}
& & & \metrichead{Wiki.}{ppl$\downarrow$} & \metrichead{LMB.}{ppl$\downarrow$}
& \avgcell{\metrichead{Avg}{ppl$\downarrow$}}
& \metrichead{BoolQ}{acc$\uparrow$} & \metrichead{PIQA}{acc$\uparrow$}
& \metrichead{Hella.}{acc-n$\uparrow$} & \metrichead{Wino.}{acc$\uparrow$}
& \metrichead{ARC-e}{acc$\uparrow$} & \metrichead{ARC-c}{acc-n$\uparrow$}
& \metrichead{OBQA}{acc-n$\uparrow$} & \metrichead{SIQA}{acc$\uparrow$}
& \avgcell{\metrichead{Avg}{acc$\uparrow$}}
& \metrichead{SWDE}{acc$\uparrow$} & \metrichead{SQuAD}{acc$\uparrow$}
& \metrichead{FDA}{acc$\uparrow$} & \metrichead{TQA}{acc$\uparrow$}
& \metrichead{NQ}{acc$\uparrow$} & \metrichead{DROP}{acc$\uparrow$}
& \avgcell{\metrichead{Avg}{acc$\uparrow$}} \\
\midrule
\multicolumn{22}{@{}l}{\textit{From-scratch training, 418M}} \\
Full attention & 418M & 2.5557 & 20.40 & 25.60 & \avgcell{23.00} & 53.64 & 68.93 & 47.57 & 55.01 & 64.35 & 32.00 & 32.40 & 39.00 & \avgcell{49.11} & 66.16 & 40.95 & 60.25 & 56.64 & 24.33 & 22.57 & \avgcell{45.15} \\
\cmidrule(lr){1-22}
NSA & 420M & 2.5727 & 21.46 & 26.13 & \avgcell{23.80} & 60.43 & 68.82 & 47.04 & 53.51 & 63.97 & 31.48 & 33.80 & 40.38 & \avgcell{49.93} & 48.51 & 38.40 & 41.11 & 55.33 & 22.84 & 18.59 & \avgcell{37.46} \\
HiLS & 423M & \textbf{2.5596} & 20.83 & 26.67 & \avgcell{23.75} & 58.99 & 70.67 & 47.28 & 54.14 & 65.11 & 31.48 & 35.00 & 40.69 & \avgcell{50.42} & 52.03 & 39.01 & 53.27 & 55.98 & 23.09 & 21.37 & \avgcell{40.79} \\
\cmidrule(lr){1-22}
\flatbsa      & 418M & 2.5684 & 21.47 & 25.02 & \avgcell{23.25} & 57.92 & 69.59 & 47.65 & 53.35 & 65.32 & 29.78 & 34.60 & 38.79 & \avgcell{49.63} & 52.30 & 40.48 & 37.57 & 53.97 & 23.34 & 21.42 & \avgcell{38.18} \\
\meanrouter   & 418M & 2.5683 & 21.53 & 24.94 & \avgcell{23.23} & 55.38 & 68.72 & 47.85 & 54.46 & 64.23 & 31.91 & 33.00 & 39.30 & \avgcell{49.36} & 52.30 & 39.11 & 45.64 & 55.39 & 21.79 & 21.18 & \avgcell{39.24} \\
\secondrouter & 418M & 2.5663 & 21.31 & 24.78 & \avgcell{23.04} & 53.12 & 69.26 & 47.32 & 55.01 & 65.66 & 30.29 & 37.60 & 39.66 & \avgcell{49.74} & 53.11 & 39.21 & 54.45 & 54.92 & 23.95 & 20.17 & \avgcell{40.97} \\
\ours         & 418M & 2.5630 & 21.17 & 24.91 & \avgcell{\textbf{23.04}} & 58.41 & 69.80 & 47.52 & 55.25 & 65.19 & 32.51 & 36.00 & 39.51 & \avgcell{\textbf{50.52}} & 57.07 & 38.81 & 54.81 & 56.16 & 23.79 & 19.98 & \avgcell{\textbf{41.77}} \\
\midrule
\multicolumn{22}{@{}l}{\textit{From-scratch training, 1.47B}} \\
Full attention & 1.47B & 2.3155 & 15.33 & 13.31 & \avgcell{14.32} & 56.33 & 73.39 & 58.37 & 58.56 & 71.89 & 38.48 & 40.80 & 42.48 & \avgcell{55.04} & 75.97 & 43.80 & 74.68 & 63.86 & 30.54 & 23.91 & \avgcell{52.13} \\
\cmidrule(lr){1-22}
NSA & 1.47B & 2.3142 & 15.70 & 11.93 & \avgcell{\textbf{13.82}} & 62.60 & 74.05 & 58.98 & 60.14 & 73.82 & 38.65 & 42.40 & 41.91 & \avgcell{\textbf{56.57}} & 68.68 & 42.29 & 59.71 & 62.74 & 28.25 & 22.14 & \avgcell{47.30} \\
HiLS & 1.48B & \textbf{2.3084} & 15.42 & 12.72 & \avgcell{14.07} & 60.31 & 73.99 & 58.79 & 59.27 & 73.32 & 39.76 & 39.60 & 41.61 & \avgcell{55.83} & 68.50 & 44.77 & 65.06 & 63.15 & 27.78 & 21.99 & \avgcell{48.54} \\
\cmidrule(lr){1-22}
\flatbsa      & 1.47B & 2.3179 & 15.63 & 12.95 & \avgcell{14.29} & 58.26 & 74.54 & 59.30 & 59.59 & 72.35 & 37.97 & 41.60 & 41.04 & \avgcell{55.58} & 68.32 & 43.10 & 64.88 & 63.09 & 27.65 & 24.20 & \avgcell{48.54} \\
\meanrouter   & 1.47B & 2.3215 & 15.77 & 13.38 & \avgcell{14.58} & 60.95 & 73.72 & 58.46 & 58.72 & 72.60 & 39.08 & 40.20 & 41.10 & \avgcell{55.60} & 67.96 & 43.90 & 56.53 & 63.51 & 29.27 & 23.24 & \avgcell{47.40} \\
\secondrouter & 1.47B & 2.3168 & 15.59 & 12.90 & \avgcell{14.24} & 62.97 & 73.39 & 58.84 & 57.06 & 71.25 & 38.23 & 39.20 & 40.48 & \avgcell{55.18} & 69.67 & 45.04 & 67.42 & 63.03 & 29.24 & 24.34 & \avgcell{49.79} \\
\ours         & 1.47B & 2.3162 & 15.58 & 13.27 & \avgcell{14.43} & 62.94 & 73.83 & 58.91 & 58.33 & 72.14 & 40.19 & 40.00 & 41.35 & \avgcell{55.96} & 71.83 & 43.93 & 69.24 & 63.27 & 29.17 & 21.56 & \avgcell{\textbf{49.83}} \\
\midrule
\multicolumn{22}{@{}l}{\textit{From-scratch training, 2.67B}} \\
Full attention & 2.67B & 2.2209 & 13.74 & 10.81 & \avgcell{12.28} & 64.50 & 74.86 & 62.37 & 62.12 & 74.92 & 43.77 & 42.00 & 41.30 & \avgcell{58.23} & 77.23 & 46.01 & 75.23 & 64.75 & 30.92 & 23.91 & \avgcell{53.01} \\
\cmidrule(lr){1-22}
NSA & 2.68B & 2.2098 & 13.76 & 9.89 & \avgcell{\textbf{11.83}} & 62.91 & 74.86 & 63.38 & 61.88 & 75.21 & 43.94 & 42.80 & 42.73 & \avgcell{58.46} & 74.35 & 45.54 & 61.07 & 66.05 & 30.73 & 23.48 & \avgcell{50.20} \\
HiLS & 2.70B & \textbf{2.2047} & 13.67 & 10.19 & \avgcell{11.93} & 62.20 & 74.92 & 63.41 & 60.06 & 76.09 & 43.60 & 42.20 & 42.53 & \avgcell{58.13} & 70.75 & 45.17 & 73.41 & 66.29 & 30.85 & 23.24 & \avgcell{51.62} \\
\cmidrule(lr){1-22}
\flatbsa      & 2.67B & 2.2184 & 13.87 & 10.34 & \avgcell{12.11} & 64.92 & 74.76 & 63.42 & 61.48 & 76.14 & 41.81 & 42.60 & 43.71 & \avgcell{\textbf{58.61}} & 72.46 & 45.64 & 67.24 & 66.53 & 30.16 & 24.72 & \avgcell{51.13} \\
\meanrouter   & 2.67B & 2.2179 & 13.96 & 10.36 & \avgcell{12.16} & 62.75 & 74.37 & 63.24 & 62.12 & 76.09 & 41.98 & 42.00 & 42.58 & \avgcell{58.14} & 70.66 & 44.81 & 68.06 & 64.87 & 30.73 & 26.11 & \avgcell{50.87} \\
\secondrouter & 2.67B & 2.2152 & 13.88 & 10.30 & \avgcell{12.09} & 62.08 & 74.81 & 63.29 & 60.62 & 76.18 & 42.75 & 42.00 & 42.94 & \avgcell{58.08} & 72.01 & 46.62 & 72.87 & 65.70 & 30.98 & 24.10 & \avgcell{52.05} \\
\ours         & 2.67B & 2.2151 & 13.83 & 10.36 & \avgcell{12.10} & 61.25 & 75.19 & 63.23 & 60.85 & 75.51 & 43.94 & 42.20 & 42.63 & \avgcell{58.10} & 75.07 & 45.95 & 69.78 & 65.58 & 31.42 & 25.06 & \avgcell{\textbf{52.14}} \\
\bottomrule
\end{tabular}%
}
\end{table*}

\subsection{Long-context evaluation}
\label{sec:long_context_evaluation}
\label{sec:cpt_results}

We evaluate long-context retrieval on four needle-in-a-haystack task families
from RULER using the models after continued pretraining at 16K.
Table~\ref{tab:cpt30_ruler} reports accuracy for the 2.67B models at
context lengths from 1K to 16K. \ours achieves strong long-context
retrieval performance.
Training loss and downstream results are provided in
Appendix~\ref{app:cpt_results}.

\begin{table*}[t]
\caption{RULER needle-in-a-haystack results after 10B tokens of continued
pretraining at 16K for the 2.67B models. Sparse methods use $C=64$ and
$K=32$. Scores are accuracies in percent. The gray Avg column averages
scores over the four task families at 1K, 2K, 4K, 8K, and 16K, before rounding.
Boldface marks the best sparse result in each column.}
\label{tab:cpt10_ruler_k32}
\label{tab:cpt30_ruler}
\centering
\scriptsize
\setlength{\tabcolsep}{1.5pt}
\resizebox{\textwidth}{!}{%
\begin{tabular}{@{}ll|rrrrr|rrrrr|rrrrr|rrrrr|r@{}}
\toprule
\multirow{2}{*}{Size} & \multirow{2}{*}{Method}
& \multicolumn{5}{c|}{\texttt{niah\_single}}
& \multicolumn{5}{c|}{\texttt{niah\_multikey}}
& \multicolumn{5}{c|}{\texttt{niah\_multiquery}}
& \multicolumn{5}{c|}{\texttt{niah\_multivalue}}
& \multirow{2}{*}{\avgcell{Avg $\uparrow$}} \\
\cmidrule(lr){3-7}\cmidrule(lr){8-12}\cmidrule(lr){13-17}\cmidrule(lr){18-22}
 & & 1K & 2K & 4K & 8K & 16K & 1K & 2K & 4K & 8K & 16K & 1K & 2K & 4K & 8K & 16K & 1K & 2K & 4K & 8K & 16K & \\
\midrule
\multirow{6}{*}[-2ex]{2.67B} & Full Attention & 99.67 & 99.60 & 99.47 & 99.13 & 98.80 & 43.33 & 38.40 & 34.67 & 28.00 & 23.13 & 85.60 & 72.45 & 50.05 & 51.55 & 34.85 & 90.15 & 84.55 & 77.05 & 57.10 & 57.20 & \avgcell{66.24} \\
\cmidrule(lr){2-23}
 & NSA & \textbf{99.93} & 99.53 & \textbf{99.33} & \textbf{97.47} & \textbf{91.67} & 57.47 & 52.00 & \textbf{40.67} & \textbf{18.00} & \textbf{15.20} & 72.60 & 79.85 & 32.65 & 50.10 & 24.20 & 83.80 & 69.55 & 59.85 & 48.65 & \textbf{28.90} & \avgcell{61.07} \\
 & \flatbsa & 99.20 & 98.40 & 93.60 & 85.33 & 59.67 & 37.33 & 32.20 & 20.87 & 13.40 & 7.13 & 80.75 & 69.85 & 66.85 & 37.90 & 15.10 & 83.80 & 75.15 & 63.45 & 38.15 & 21.70 & \avgcell{54.99} \\
 & \meanrouter & 99.87 & \textbf{99.73} & 97.33 & 81.87 & 60.27 & 66.80 & \textbf{53.40} & 39.40 & 12.60 & 9.93 & \textbf{93.40} & \textbf{89.90} & 73.75 & 38.50 & 21.70 & 83.30 & 82.95 & \textbf{70.50} & 38.15 & 20.45 & \avgcell{61.69} \\
 & \secondrouter & 99.47 & 99.33 & 99.00 & 90.00 & 72.40 & 51.13 & 40.00 & 27.20 & 13.13 & 9.33 & 91.45 & 86.95 & 77.85 & \textbf{59.55} & \textbf{25.20} & 89.60 & 81.25 & 69.60 & \textbf{52.55} & 23.30 & \avgcell{\textbf{62.92}} \\
 & \ours & 99.07 & 94.60 & 96.33 & 91.47 & 71.27 & \textbf{68.20} & 52.33 & 32.47 & 13.53 & 10.00 & 88.65 & 85.35 & \textbf{79.00} & 48.50 & 20.85 & \textbf{93.40} & \textbf{87.10} & 58.60 & 43.95 & 21.30 & \avgcell{62.80} \\
\bottomrule
\end{tabular}%
}
\end{table*}

\subsection{Ablation Studies}
\label{sec:ablations}

\paragraph{Scoring function.}
\label{sec:scoring_ablation}
We compare \ours with \meanrouter and \secondrouter, keeping the backbone
and hierarchy unchanged. For logits $z_1,\ldots,z_m$ with mean
$\bar z$ and variance $\operatorname{Var}(z)$, Taylor expansion gives
\begin{equation}
\begin{aligned}
 s_{\mathrm{PISA}}&=\log\sum_{j=1}^{m}e^{z_j}\\
 &=\log m+\bar z+\frac12\operatorname{Var}(z)
 +\text{higher-order terms}.
\end{aligned}
\label{eq:lse_approximations}
\end{equation}
Omitting $\log m$, \meanrouter and \secondrouter truncate this expansion
at first and second order, respectively:
\begin{align}
 s_{\mathrm{PISA\text{-}1}}&=\bar z,\\
 s_{\mathrm{PISA\text{-}2}}&=\bar z+\frac12\operatorname{Var}(z).
\end{align}
The pretraining results in Table~\ref{tab:native_main} show that \ours has lower
training loss and higher average containment accuracy than \meanrouter,
\secondrouter, and BSA at all three scales.
Appendix~\ref{app:scoring_details} provides the scoring derivations.

\Needspace{14\baselineskip}
\paragraph{Block-selection quality.}
\label{sec:oracle_results}
We evaluate all block selectors on identical query and key tensors from
a Full-Attention checkpoint. For each query position $i$ and KV head $h$,
each method selects $K$ blocks using its scoring rule, and we construct
a reference set of $K$ blocks from the full-attention weights.
We denote these sets by $I_{\mathrm{sparse}}^{i,h}$ and
$I_{\mathrm{full}}^{i,h}$, respectively.
Let $P_{i,h}(B)$ denote the full-attention mass assigned to block $B$,
averaged over the query heads sharing KV head $h$.
We report two metrics:
\begingroup
\begin{align}
 \mathrm{Recall@K}_{i,h}
 &=\frac{|I_{\mathrm{sparse}}^{i,h}\cap I_{\mathrm{full}}^{i,h}|}{K},\\
 \mathrm{MassRatio}_{i,h}
 &=\frac{\sum_{B\in I_{\mathrm{sparse}}^{i,h}}P_{i,h}(B)}
         {\sum_{B\in I_{\mathrm{full}}^{i,h}}P_{i,h}(B)}.
\label{eq:oracle_metrics}
\end{align}
\endgroup
Recall@K measures the overlap between the selected and reference blocks;
attention mass ratio measures the attention mass retained by the selected
blocks relative to the reference set, not total full-attention mass.
In this experiment, \ours achieves the highest average Recall@8 and
attention mass ratio (Figure~\ref{fig:oracle_by_position}).
The complete evaluation setup and aggregate results are provided in
Appendix~\ref{app:oracle_details}. This indicates that LSE-based scoring enables more accurate Top-\(K\) selection.

\begin{figure}[!htbp]
\centering
\includegraphics[width=\linewidth]{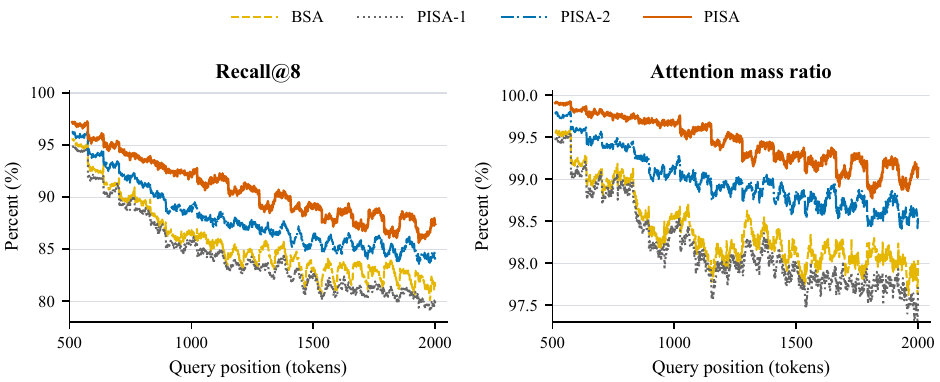}
\caption{Block-selection quality on 100 FDA prompts using identical
Full-Attention queries and keys. Curves show mean Recall@8 and attention
mass ratio by query position. All selectors and the reference share $K=8$
and the forced-block policy; Recall@8 includes forced blocks.}
\label{fig:oracle_by_position}
\end{figure}

\Needspace{0.65\textheight}
\subsection{Block-selection efficiency}
\label{sec:routing_efficiency}

\begin{figure}[!htbp]
\centering
\includegraphics[width=0.9\linewidth]{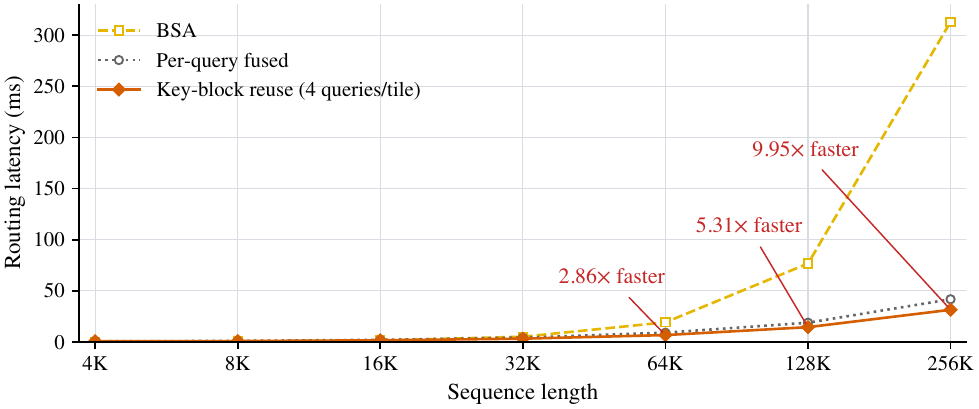}
\caption{Prefill block-selection latency from 4K to 256K with $C=64$ and $K=8$. Timings
include mean-summary construction and all selection stages, but exclude
attention over the selected blocks. The key-block reuse implementation of
\ours uses $Q_{\mathrm{tile}}=4$; annotations show speedups over BSA. Complete settings are provided
in Appendix~\ref{app:kernel_microbenchmark}.}
\label{fig:routing_latency_scaling}
\end{figure}

Figure~\ref{fig:routing_latency_scaling} compares prefill block-selection
latency for BSA and the Q4 implementation of \ours ($Q_{\mathrm{tile}}=4$).
BSA is faster from 4K to 16K, whereas \ours achieves $2.86\times$,
$5.31\times$, and $9.95\times$ speedups over BSA at 64K, 128K, and 256K, respectively.
At the same lengths, Q4 achieves $1.35\times$, $1.30\times$, and $1.33\times$
speedups over the per-query implementation, respectively. Detailed timings
are provided in
Appendix~\ref{app:kernel_microbenchmark}
(Table~\ref{tab:kernel_grouped_benchmark}).

\FloatBarrier
\section{Conclusion}
\label{sec:conclusion}

We introduced \ours, a block-sparse attention method that combines pyramid
Top-$K$ selection with LogSumExp scoring. By scoring a bounded candidate set
at each level, \ours reduces prefill complexity to $O(N\log N)$ for fixed
model dimensions and selection parameters. Our Triton kernels fuse
intermediate selection levels and reuse leaf-key blocks across queries.
Experiments across three model scales show comparable language-modeling and
commonsense-reasoning performance to sparse baselines, with higher average
containment accuracy after pretraining. Block-selection benchmarks further
demonstrate lower latency than BSA at long sequence lengths.

\newpage
\bibliography{references}
\bibliographystyle{plainnat}

\clearpage
\appendix
\makeatletter
\newcommand{\appendixcontents}{%
  \section*{Appendix Contents}%
  \label{app:roadmap}%
  \begingroup
  \setlength{\parskip}{3pt}%
  \def\l@appendixsection##1##2{\par\noindent ##1 \dotfill ##2\par}%
  \renewcommand\numberline[1]{\textbf{Appendix~##1}\quad}%
  \@starttoc{apc}%
  \endgroup
}
\newcommand{\appsection}[1]{%
  \section{#1}%
  \addcontentsline{apc}{appendixsection}{\protect\numberline{\thesection}#1}%
}
\makeatother

\appendixcontents

\appsection{Analysis of Block Scoring}
\label{app:derivations}
\label{app:scoring_details}

For a fixed query head, a visible block's full-attention mass is
proportional to the exponential of its raw-key LSE score, so both give
the same block ranking. Intermediate blocks are instead scored from
child mean summaries, which do not retain all raw-key information.

\subsection{Intermediate block scoring}
\label{app:jensen_bound}

At intermediate levels, \ours avoids reading all keys in a parent block
by scoring its child mean summaries. We compare LSE over these summaries
with averaging their logits, using the raw-key LSE as the reference.

Fix a query $q$ and one query head. Let $B$ be a fully visible block
with $g$ equally sized children $C_1,\ldots,C_g$. Assume that every
preceding pooling step also combines equally sized groups, so each
stored child summary equals the mean of its original keys.
Writing $z_t=q^\top k_t/\sqrt d$, define the child mean logits and
the parent's normalized raw-key LSE as
\begin{align}
 m_j&=\frac1{|C_j|}\sum_{t\in C_j}z_t,\\
 F_q(B)&=\log\left(\frac1{|B|}\sum_{t\in B}e^{z_t}\right).
\end{align}
The scaled dot product between $q$ and child $j$'s mean summary equals
$m_j$. Under these conditions,
\begin{equation}
 \frac1g\sum_{j=1}^g m_j
 \;\le\;
 \widehat F(B):=\log\left(\frac1g\sum_{j=1}^g e^{m_j}\right)
 \;\le\;
 F_q(B).
 \label{eq:equal_child_jensen}
\end{equation}
The left term is the intermediate mean score used by \meanrouter.
The middle term is the \ours score minus $\log g$, and the right
term is the parent raw-key LSE minus $\log|B|$.
These normalizations remove the input-count terms so that the scores
can be compared on the same scale; they do not change rankings among
candidates with the same input counts.

For the left inequality in Equation~\ref{eq:equal_child_jensen},
the convexity of the exponential function allows Jensen's inequality
to be applied to the $g$ child mean logits:
\begin{equation}
 \exp\left(\frac1g\sum_{j=1}^g m_j\right)
 \le \frac1g\sum_{j=1}^g e^{m_j}.
\end{equation}
Since the logarithm is increasing, taking logarithms on both sides gives
$g^{-1}\sum_jm_j\le\widehat F(B)$.

For the right inequality, apply Jensen separately to the raw-key logits
within each child $C_j$:
\begin{equation}
 e^{m_j}
 =\exp\left(\frac1{|C_j|}\sum_{t\in C_j}z_t\right)
 \le\frac1{|C_j|}\sum_{t\in C_j}e^{z_t}.
\end{equation}
The children partition $B$ and each has $|C_j|=|B|/g$ tokens.
Averaging the inequalities over the $g$ children therefore gives
\begin{equation}
 \frac1g\sum_{j=1}^g e^{m_j}
 \le\frac1g\sum_{j=1}^g
       \frac1{|C_j|}\sum_{t\in C_j}e^{z_t}
 =\frac1{|B|}\sum_{t\in B}e^{z_t}.
\end{equation}
Taking logarithms gives $\widehat F(B)\le F_q(B)$, completing the
two inequalities.

Thus, for the same block, normalized LSE over child means is at least
as close to the normalized raw-key LSE as the mean score is.
It uses differences among child means, but cannot recover variation
within each child. When all logits within each child are equal,
$\widehat F(B)=F_q(B)$. This comparison concerns the numerical error
of a block score; it does not guarantee more accurate block rankings
or recovery of the globally highest-mass leaf blocks.

\subsection{First- and second-order approximations}
\label{app:lse2_details}

\meanrouter uses the first-order Taylor approximation of normalized
LSE, while \secondrouter also retains the second-order term.
For one candidate and query head, let $z_1,\ldots,z_m$ be the valid
logits entering its score, with $m\ge1$. Define their mean and
population variance by
\begin{equation}
 \bar z=\frac1m\sum_{j=1}^m z_j,
\end{equation}
\begin{equation}
\begin{aligned}
 v_z&=\operatorname{Var}(z)\\
 &=\frac1m\sum_{j=1}^m(z_j-\bar z)^2.
\end{aligned}
 \label{eq:direct_score_statistics}
\end{equation}
To derive the approximations, introduce
\begin{equation}
 \phi(\lambda)=\log\left(\frac1m\sum_{j=1}^m e^{\lambda z_j}\right).
 \label{eq:block_cgf}
\end{equation}
Here, $\phi(1)=s_{\mathrm{PISA}}-\log m$ is normalized LSE. Direct differentiation gives
\begin{align}
 \phi(0)&=0,\\
 \phi'(0)&=\bar z,\\
 \phi''(0)&=v_z.
\end{align}
Taylor's theorem about $\lambda=0$, evaluated at $\lambda=1$, gives
\begin{equation}
 \phi(1)=\bar z+\frac12v_z+
 \frac12\int_0^1(1-\lambda)^2\phi'''(\lambda)\,d\lambda.
 \label{eq:exact_second_remainder}
\end{equation}
Keeping only the mean gives \meanrouter; adding half the variance
gives \secondrouter. In contrast, \ours computes LSE without
truncating the expansion:
\begin{align}
 s_{\mathrm{PISA}}&=\log\sum_{j=1}^{m}e^{z_j},\\
 s_{\mathrm{PISA\text{-}1}}&=\bar z,\\
 s_{\mathrm{PISA\text{-}2}}&=\bar z+\tfrac12v_z.
 \label{eq:three_scores_one_support}
\end{align}

\subsection{Grouped-head scoring}
\label{app:gqa_objective}

In grouped-query attention~\citep{ainslie2023gqa}, \ours selects one
shared block set for query heads that share a KV head. For a fixed
query position, let $\mathcal H(h)$ denote the query heads sharing
KV head $h$, with $G_Q=|\mathcal H(h)|$. \ours sums their LSE scores:
\begin{equation}
 U_B=\sum_{h'\in\mathcal H(h)}s_{h',B},
 \label{eq:gqa_product_objective}
\end{equation}
where $s_{h',B}$ is the score of block $B$ for query head $h'$.
The full-attention reference in the diagnostic instead averages their block masses:
\begin{equation}
 P_h(B)=\frac1{G_Q}\sum_{h'\in\mathcal H(h)}p_{h'}(B),
 \label{eq:gqa_normalized_mass_objective}
\end{equation}
where $p_{h'}(B)$ is the total attention probability assigned to
block $B$ by head $h'$. Even with exact raw-key LSE scores, these
two aggregation rules can rank blocks differently.

\appsection{Implementation Details}
\label{app:kernel_details}

\paragraph{Causal masking.}
\label{app:kernel_causality}
\label{app:causal_ablation}
Candidates outside the sequence or starting after the query are
excluded. We retain the first, previous, and current leaf blocks and
their ancestor paths whenever distinct and eligible, following the
NSA implementation in Flash Linear Attention~\citep{yang2024fla}.
Prefill summaries on the current block's path can contain future keys,
but these nodes are retained without using their scores for ranking.
All other eligible nodes lie entirely in the query's past, so future
keys cannot affect selection. The attention kernel separately masks
future tokens within selected blocks.

During backpropagation, the selected block indices are held fixed:
gradients flow through attention on the selected entries, not through
the discrete selection decisions.

\paragraph{Decode cache.}
\label{app:incremental_decode}
The cached pyramid stores $O((N/C)d)$ elements per KV head, and each
level's capacity grows by a fixed fraction when more space is needed.
Each level therefore expands only a constant number of times as the
sequence length doubles; even if every expansion copies the entire
summary buffer, the total copying cost through length $N$ is
$O((N/C)d\log N)$ per KV head.
At fixed model and selection parameters, these occasional expansions
preserve the $O(\log N)$ average decoding cost per generated token.

\begin{center}
\begin{minipage}{\textwidth}
\centering
\small
\captionof{table}{Fixed-length block-selection latency from 4K to 256K.
Q1, Q2, and Q4 reuse one leaf-key tile across up to one, two, and four query
references, respectively. Boldface marks the lowest latency at each sequence length.}
\label{tab:kernel_grouped_benchmark}
\footnotesize
\setlength{\tabcolsep}{3.4pt}
\renewcommand{\arraystretch}{1.12}
\begin{tabular}{@{}rrrrrr@{}}
\toprule
\multicolumn{6}{c}{\textbf{Block-selection latency (ms)}} \\
\midrule
$N$ & Per-query & Q1 & Q2 & Q4 & BSA \\
\midrule
4K   & 0.573228 & 0.716264 & 0.702543 & 0.701181 & \textbf{0.167007} \\
8K   & 1.028081 & 1.175723 & 0.908852 & 0.864647 & \textbf{0.462596} \\
16K  & 2.049632 & 2.295943 & 1.696250 & 1.581351 & \textbf{1.461185} \\
32K  & 4.243254 & 4.684515 & 3.429051 & \textbf{3.184846} & 5.098899 \\
64K  & 9.040538 & 9.716966 & 7.183495 & \textbf{6.704480} & 19.143526 \\
128K & 18.813094 & 20.466775 & 15.466555 & \textbf{14.452411} & 76.681923 \\
256K & 41.732992 & 43.365215 & 33.521616 & \textbf{31.440747} & 312.955505 \\
\bottomrule
\end{tabular}

\end{minipage}
\end{center}

\paragraph{Block-selection benchmarks.}
\label{app:kernel_microbenchmark}

Table~\ref{tab:kernel_grouped_benchmark} reports fixed-length
block-selection timings for Figure~\ref{fig:routing_latency_scaling}.
We use random BF16 queries and keys with a fixed seed, batch size 1,
32 query heads, two KV heads, $d=64$, $C=64$, $K=8$, and $g=2$,
retaining the first, previous, and current blocks whenever eligible.
Dot products accumulate in FP32, as do score reductions.
The per-query variant fuses all selection levels per query and KV head;
Q1, Q2, and Q4 instead use separate leaf-scoring programs with one,
two, and four query slots, respectively.
Timings use Triton's benchmarking utility after warm-up and include
summary construction and all selection stages, but exclude sparse attention.

Q4 is slower than per-query at 4K but faster from 8K onward,
reaching $1.35\times$, $1.30\times$, and $1.33\times$ speedups at 64K, 128K, and 256K,
respectively. BSA is fastest through 16K; Q4 has the lowest reported latency
from 32K to 256K.

\appsection{Experimental setup}
\label{app:setup}

\subsection{Model and training configuration}

All models use the GPT-2 BPE tokenizer~\citep{radford2019language} with
a vocabulary of 50{,}257 tokens; the embedding and output matrices are padded to
50{,}432 rows. Decoder blocks use pre-RMSNorm, SiLU-gated feed-forward
layers, and no dropout. RoPE is applied to the high-frequency half
of each attention head with a base of 10{,}000.
Table~\ref{tab:training_config} summarizes the architecture dimensions
and from-scratch training settings.

\begin{table}[h]
\caption{From-scratch training configuration.}
\label{tab:training_config}
\centering
\small
\begin{tabular}{p{\dimexpr0.34\linewidth-2\tabcolsep\relax}*{3}{>{\centering\arraybackslash}p{\dimexpr0.22\linewidth-2\tabcolsep\relax}}}
\toprule
Field & 418M & 1.47B & 2.67B \\
\midrule
Layers & 24 & 24 & 32 \\
Model dimension & 1024 & 2048 & 2560 \\
MLP dimension & 2816 & 5632 & 6912 \\
Query heads / KV heads & 32 / 2 & 32 / 2 & 32 / 2 \\
Head dimension & 64 & 128 & 128 \\
Sequence length & 4096 & 4096 & 4096 \\
Per-device batch / accumulation & 8 / 2 & 4 / 2 & 4 / 2 \\
Training steps / warm-up & 100K / 1K & 100K / 1K & 100K / 1K \\
Training budget (B tokens) & 100 & 100 & 100 \\
Peak LR & $3\times10^{-4}$ & $3\times10^{-4}$ & $3\times10^{-4}$ \\
Seed & 42 & 42 & 42 \\
Optimizer & \multicolumn{3}{>{\centering\arraybackslash}p{\dimexpr0.66\linewidth-2\tabcolsep\relax}}{AdamW, $\beta=(0.9,0.95)$, $\epsilon=10^{-8}$, weight decay $0.1$} \\
LR schedule & \multicolumn{3}{>{\centering\arraybackslash}p{\dimexpr0.66\linewidth-2\tabcolsep\relax}}{1K linear warm-up; 90K plateau; 9K square-root decay to $0.1\times$ peak} \\
Gradient clipping & \multicolumn{3}{>{\centering\arraybackslash}p{\dimexpr0.66\linewidth-2\tabcolsep\relax}}{global norm $1.0$} \\
FSDP precision & \multicolumn{3}{>{\centering\arraybackslash}p{\dimexpr0.66\linewidth-2\tabcolsep\relax}}{bfloat16 parameters, FP32 reductions} \\
\bottomrule
\end{tabular}
\end{table}

At each model scale, the training data, decoder backbone, and attention
projections are matched across methods. Sparse methods use $C=64$ and
$K=8$. We follow the default policy in the NSA implementation of Flash
Linear Attention~\citep{yang2024fla}, retaining the first, previous,
and current blocks whenever eligible.
\ours, \meanrouter, and \secondrouter share the same hierarchy,
candidate-expansion rule, and selected-attention code; they differ only
in the block score.
\flatbsa uses NSA's selected-attention branch with non-hierarchical
block selection. Unlike \ours and its variants, it normalizes candidate
scores within each query head before summing across heads.

\paragraph{Continued pretraining.}
We use $C=64$ and $K=32$ for sparse methods during 10B tokens of
continued pretraining at 16K. Each run uses
the same block budget during training and evaluation.

Each continued-pretraining run starts from the corresponding
4K pretrained checkpoint, restoring model parameters and Adam optimizer
states. We extend the sequence length to 16K and increase the RoPE base
from 10{,}000 to 80{,}000. Each run uses a new learning-rate schedule:
linear warm-up to $3\times10^{-5}$ over the first 10\% of training steps,
followed by cosine decay to $3\times10^{-6}$. Packed inputs preserve
document boundaries, with no cross-document attention.

\subsection{Evaluation details}

\paragraph{Language modeling and multiple-choice evaluation.}
We use the Language Model Evaluation Harness~\citep{schoelkopf2024lmeval} for
WikiText word perplexity~\citep{merity2016wikitext}, LAMBADA
perplexity~\citep{paperno2016lambada}, and eight multiple-choice tasks:
BoolQ, PIQA, HellaSwag, WinoGrande, ARC-Easy, ARC-Challenge, OpenBookQA,
and SocialIQA~\citep{clark2019boolq,bisk2020piqa,zellers2019hellaswag,
sakaguchi2020winogrande,clark2018arc,mihaylov2018openbookqa,sap2019socialiqa}.
WikiText and LAMBADA use the standard \texttt{wikitext} and
\texttt{lambada\_openai} tasks; their reported average is the arithmetic
mean of the two perplexities. For multiple-choice evaluation, we report
acc-n for HellaSwag, ARC-Challenge, and OpenBookQA, and acc for the other
five tasks, as indicated in Table~\ref{tab:native_main}.
The multiple-choice average (Acc-8) gives equal weight to these eight
task scores. These tasks and metrics are unchanged after continued pretraining.

\paragraph{Containment evaluation.}
We evaluate six retrieval tasks: SWDE, SQuAD Completion, FDA,
TriviaQA~\citep{joshi2017triviaqa}, Natural
Questions~\citep{kwiatkowski2019natural}, and DROP~\citep{dua2019drop}.
We use the task implementations from BASED and
JRT~\citep{arora2024based,arora2024jrt}.
Contains counts a prediction as correct if it includes any gold answer
as a case-insensitive literal substring. All six tasks use zero-shot
greedy decoding on the validation split, generating at most 48 tokens
and stopping at a newline or EOS.

\paragraph{RULER evaluation.}
We evaluate the continued-pretraining models on four needle-in-a-haystack
families from RULER~\citep{hsieh2024ruler}: single-key, multi-key,
multi-query, and multi-value. We use greedy decoding at context lengths
of 1K, 2K, 4K, 8K, and 16K, as reported in Table~\ref{tab:cpt30_ruler}.
The single-key and multi-key scores each
average three task templates, while multi-query and multi-value each
use one template. Avg is the mean over the four families and five
context lengths, computed before rounding.
\FloatBarrier
 
\appsection{Additional Experiments}
\label{app:additional_results}

\subsection{Downstream evaluation after continued pretraining}
\label{app:cpt_results}

Table~\ref{tab:cpt30_downstream} reports language modeling and downstream
performance after 10B tokens of continued pretraining at 16K. \ours has
slightly lower training loss than BSA and \meanrouter
at both scales, with perplexity and average multiple-choice accuracy
comparable to BSA. Its average containment accuracy is lower than BSA
at both scales.

\begin{table*}[!htb]
\caption{Training loss and downstream performance for the 1.47B and 2.67B models
after 10B tokens of continued pretraining at 16K. Sparse methods,
including NSA~\citep{yuan2025nsa}, use
$C=64$ and $K=32$.
Loss is the final training loss. Gray columns show group averages;
the containment average covers SWDE, SQuAD Completion, and FDA.
The block budget does not apply to Full Attention.}
\label{tab:cpt10_downstream}
\label{tab:cpt30_downstream}
\centering
\scriptsize
\setlength{\tabcolsep}{1.5pt}
\resizebox{\textwidth}{!}{%
\begin{tabular}{@{}lcc|c|ccc|ccccccccc|cccc@{}}
\toprule
\multirow{2}{*}[-3ex]{Method} & \multirow{2}{*}[-3ex]{Params} & \multirow{2}{*}[-3ex]{$K$} & \multirow{2}{*}[-3ex]{\metrichead{Loss}{$\downarrow$}} & \multicolumn{3}{c|}{Perplexity} & \multicolumn{9}{c|}{Multiple-choice} & \multicolumn{4}{c}{Containment} \\
\cmidrule(lr){5-7}\cmidrule(lr){8-16}\cmidrule(lr){17-20}
 &  &  &  & \metrichead{Wiki.}{ppl$\downarrow$} & \metrichead{LMB.}{ppl$\downarrow$} & \avgcell{\metrichead{Avg}{ppl$\downarrow$}} & \metrichead{BoolQ}{acc$\uparrow$} & \metrichead{PIQA}{acc$\uparrow$} & \metrichead{Hella.}{acc-n$\uparrow$} & \metrichead{Wino.}{acc$\uparrow$} & \metrichead{ARC-e}{acc$\uparrow$} & \metrichead{ARC-c}{acc-n$\uparrow$} & \metrichead{OBQA}{acc-n$\uparrow$} & \metrichead{SIQA}{acc$\uparrow$} & \avgcell{\metrichead{Avg}{acc$\uparrow$}} & \metrichead{SWDE}{acc$\uparrow$} & \metrichead{SQuAD}{acc$\uparrow$} & \metrichead{FDA}{acc$\uparrow$} & \avgcell{\metrichead{Avg}{acc$\uparrow$}} \\
\midrule
\multicolumn{20}{@{}l}{\textit{10B CPT, 1.47B}} \\
Full Attention & 1.47B & -- & 1.6993 & 13.33 & 10.04 & \avgcell{11.69} & 51.65 & 73.99 & 58.28 & 57.85 & 71.55 & 38.05 & 41.00 & 40.28 & \avgcell{54.08} & 76.87 & 46.68 & 75.14 & \avgcell{66.23} \\
\cmidrule(lr){1-20}
NSA & 1.47B & 32 & 1.6990 & 13.42 & 9.67 & \avgcell{11.55} & 63.82 & 73.61 & 58.47 & 59.67 & 72.47 & 37.97 & 42.60 & 40.23 & \avgcell{56.11} & 74.44 & 44.84 & 72.23 & \avgcell{63.84} \\
\flatbsa & 1.47B & 32 & 1.7039 & 13.36 & 9.92 & \avgcell{11.64} & 56.45 & 74.32 & 58.64 & 59.35 & 71.51 & 37.37 & 40.00 & 40.84 & \avgcell{54.81} & 76.06 & 44.54 & 79.04 & \avgcell{66.55} \\
\meanrouter & 1.47B & 32 & 1.7042 & 13.45 & 9.95 & \avgcell{11.70} & 59.57 & 73.39 & 58.32 & 58.88 & 71.04 & 37.20 & 40.00 & 40.17 & \avgcell{54.82} & 74.44 & 44.27 & 71.05 & \avgcell{63.25} \\
\secondrouter & 1.47B & 32 & 1.7007 & 13.38 & 9.81 & \avgcell{11.59} & 63.39 & 73.39 & 58.77 & 57.06 & 70.71 & 38.31 & 40.60 & 39.76 & \avgcell{55.25} & 72.19 & 46.45 & 73.32 & \avgcell{63.99} \\
\ours & 1.47B & 32 & 1.7010 & 13.37 & 9.99 & \avgcell{11.68} & 61.59 & 74.32 & 58.38 & 57.93 & 71.30 & 39.93 & 41.40 & 40.99 & \avgcell{55.73} & 74.89 & 45.41 & 75.68 & \avgcell{65.33} \\
\midrule
\multicolumn{20}{@{}l}{\textit{10B CPT, 2.67B}} \\
Full Attention & 2.67B & -- & 1.6312 & 11.96 & 8.12 & \avgcell{10.04} & 64.98 & 75.19 & 62.71 & 61.25 & 74.54 & 42.49 & 41.80 & 40.28 & \avgcell{57.91} & 77.23 & 48.19 & 75.41 & \avgcell{66.94} \\
\cmidrule(lr){1-20}
NSA & 2.68B & 32 & 1.6243 & 11.77 & 7.70 & \avgcell{9.73} & 65.90 & 75.03 & 63.31 & 60.46 & 73.57 & 41.98 & 42.00 & 41.56 & \avgcell{57.98} & 79.84 & 48.83 & 75.68 & \avgcell{68.12} \\
\flatbsa & 2.67B & 32 & 1.6324 & 11.91 & 7.99 & \avgcell{9.95} & 63.39 & 75.19 & 63.20 & 61.48 & 74.07 & 40.10 & 42.60 & 40.48 & \avgcell{57.56} & 77.68 & 47.65 & 79.13 & \avgcell{68.15} \\
\meanrouter & 2.67B & 32 & 1.6331 & 11.96 & 8.03 & \avgcell{10.00} & 63.61 & 74.97 & 63.31 & 61.96 & 74.49 & 41.30 & 42.40 & 41.30 & \avgcell{57.92} & 77.68 & 48.39 & 73.96 & \avgcell{66.68} \\
\secondrouter & 2.67B & 32 & 1.6302 & 11.90 & 7.91 & \avgcell{9.91} & 57.19 & 75.03 & 62.91 & 61.01 & 75.21 & 42.92 & 43.20 & 40.07 & \avgcell{57.19} & 75.61 & 48.76 & 80.85 & \avgcell{68.41} \\
\ours & 2.67B & 32 & 1.6297 & 11.88 & 7.95 & \avgcell{9.91} & 59.45 & 75.73 & 63.31 & 61.64 & 73.91 & 43.94 & 41.20 & 40.17 & \avgcell{57.42} & 77.50 & 49.10 & 72.60 & \avgcell{66.40} \\
\bottomrule
\end{tabular}%
}
\end{table*}

\subsection{Block-selection diagnostic}
\label{app:oracle_details}

We replay all selectors on identical position-encoded query and key
tensors from a 418M Full Attention model pretrained for 100B tokens at 4K.
The model weights are fixed, and no additional training is performed.
We use all 24 layers, $C=64$, and $K=8$. We sample 100 FDA prompts with
seed 0, restricting prompt lengths to 513--4096 tokens; the sampled
prompts contain 618--2001 tokens. We evaluate zero-based query positions
$i\ge512$, where more than eight blocks are eligible.

Section~\ref{sec:oracle_results} defines the full-attention reference set,
Recall@K, and attention mass ratio. For KV head $h$, let $\mathcal H(h)$ contain its query
heads and let $\mathcal E_i$ contain the causally available leaf blocks at
query position $i$. We define the log mass of block $B$ for query head $h'$
and its grouped mass as
\begin{align}
 a_{h',i,h}(B)
 &=\log\sum_{\substack{t\in B\\t\le i}}
   \exp\!\left(\frac{q_{i,h'}^{\top}k_{t,h}}{\sqrt d}\right),\\
 P_{i,h}(B)
 &=\frac{1}{|\mathcal H(h)|}\sum_{h'\in\mathcal H(h)}
   \frac{e^{a_{h',i,h}(B)}}
        {\sum_{B'\in\mathcal E_i}e^{a_{h',i,h}(B')}}.
\label{eq:oracle_block_mass}
\end{align}
Here, $q_{i,h'}$ and $k_{t,h}$ are the query and key vectors at positions
$i$ and $t$, respectively. Thus, $P_{i,h}(B)$ is the share of full-attention
mass assigned to $B$, averaged over the query heads that share KV head $h$.

The selector and full-attention reference use the same eligible blocks, budget, and
forced-block policy, retaining the first, previous, and current blocks.
The reference fills the remaining slots with the eligible blocks of highest full-attention mass.
Recall@8 therefore includes a guaranteed three-block overlap. The reference's
average of normalized attention masses differs from the sum of query-head
LSE scores used by \ours (Appendix~\ref{app:gqa_objective}); the two can
rank blocks differently. We also report captured mass:
\begin{equation}
 \mathrm{CapturedMass}_{i,h}
 =\sum_{B\in I_{\mathrm{sparse}}^{i,h}}P_{i,h}(B).
\end{equation}
Captured mass is the fraction of total full-attention mass retained.
Attention mass ratio divides it by the mass captured by the reference
set $I_{\mathrm{full}}^{i,h}$, not by total full-attention mass.
Aggregate scores assign equal weight to each
$(\text{prompt},\text{position},\text{layer},\text{KV head})$ decision,
so longer prompts contribute more decisions. At each query position,
Figure~\ref{fig:oracle_by_position} averages over the available prompts,
layers, and KV heads without binning or smoothing.

\begin{center}
\begin{minipage}{0.82\textwidth}
\captionof{table}{Block-selection quality on 100 FDA prompts using shared
Full-Attention queries and keys, with $C=64$ and $K=8$. Scores are percentages, averaged over query rows with more than
eight eligible blocks. Recall@8 includes the three forced blocks.
Boldface marks the best result in each column.}
\label{tab:oracle_fda_aggregate}
\centering
\small
\setlength{\tabcolsep}{4pt}
\begin{tabular}{lrrr}
\toprule
Selector & Recall@8 $\uparrow$ & Captured mass $\uparrow$ & Attention mass ratio $\uparrow$ \\
\midrule
\flatbsa & 85.91 & 85.67 & 98.42 \\
\meanrouter & 84.75 & 85.50 & 98.22 \\
\secondrouter & 88.24 & 86.13 & 99.02 \\
\ours & \textbf{90.95} & \textbf{86.47} & \textbf{99.46} \\
\bottomrule
\end{tabular}
\end{minipage}
\end{center}

Table~\ref{tab:oracle_fda_aggregate} shows that \ours has the highest
Recall@8, captured mass, and attention mass ratio in this diagnostic.
The selected blocks capture nearly as much attention mass as the
full-attention reference.

\appsection{Limitations}
\label{sec:limitations}
Computational constraints limit the model sizes and pretraining data budgets
explored in our experiments. Both factors can substantially affect performance
and relative gains over baselines. Within the evaluated settings, our central
finding remains that pyramid block selection reduces selection complexity
while maintaining competitive model performance.

\end{document}